\pdfoutput=1

\documentclass[twocolumn,11pt]{article}

\usepackage[preprint]{acl}
\usepackage{comment}
\usepackage{times}
\usepackage{latexsym}

\usepackage[T1]{fontenc}

\usepackage[utf8]{inputenc}

\usepackage{microtype}

\usepackage{inconsolata}

\usepackage{graphicx}

\usepackage{url}

\usepackage{natbib}

\usepackage{colortbl}
\definecolor{lightgreen}{rgb}{0.92, 1, 0.92}

\usepackage{booktabs}
\usepackage{multirow}
\usepackage{float}
\usepackage{caption}
\usepackage[normalem]{ulem}
\useunder{\uline}{\ul}{}

\usepackage{hyphenat}

\usepackage{fontawesome5}

\title{
    \raisebox{-0.3\height}{\includegraphics[width=0.08\textwidth]{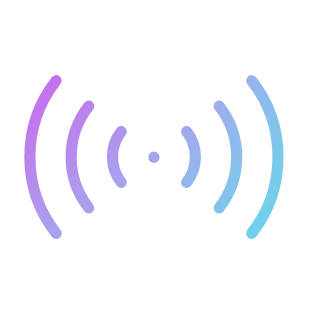}}
    SpeechT: Findings of the First Mentorship in Speech Translation}

\author{\textbf{\normalsize Yasmin Moslem\textsuperscript{\tiny\faStar[regular]}} \hspace{0.5em}
  \textbf{\normalsize Juan Julián Cea Morán*} \hspace{0.5em}
  \textbf{\normalsize Mariano Gonzalez-Gomez*} \\ \\ 
  \textbf{\normalsize Muhammad Hazim Al Farouq*} \hspace{0.5em} 
  \textbf{\normalsize Farah Abdou*} \hspace{0.5em}
  \textbf{\normalsize Satarupa Deb*}
  }

\date{}

\begin{document}
\maketitle
\def\thefootnote{{\scalebox{0.5}{\faStar[regular]}}}
\footnotetext{Correspondence: \url{yasmin [at] machinetranslation.io}}\def\thefootnote{\arabic{footnote}}
\def\thefootnote{*}\footnotetext{Participant in the mentorship}\def\thefootnote{\arabic{footnote}}

\begin{abstract}
\nohyphens{
This work presents the details and findings of the first mentorship in speech translation (SpeechT), which took place in December 2024 and January 2025. To fulfil the mentorship requirements, the participants engaged in key activities, including data preparation, modelling, and advanced research. The participants explored data augmentation techniques and compared end-to-end and cascaded speech translation systems. The projects covered various languages other than English, including Arabic, Bengali, Galician, Indonesian, Japanese, and Spanish.
}

\end{abstract}

\section{Introduction}

At the beginning of the mentorship on speech translation, the participants were provided with the following descriptions and guidelines for each task:

\paragraph{Data:} Define, collect, and process bilingual speech data in a chosen language. Your dataset should consist of “train”, “dev/validation”, and “test” splits. By the end of the task, each participant should share a Hugging Face link to their datasets. The dataset page metadata should include sections for data sources, processing steps you applied in detail, and credits/citations of the original datasets. 

\paragraph{Modelling:} Choose one of the popular models, e.g. Whisper \citep{Radford2022-Whisper} or Wav2Vec \citep{Baevski2020-Wav2Vec}, and fine-tune it on the data prepared in the first task. Experimenting with different fine-tuning approaches and hyperparameters is encouraged. By the end of the task, the participants should share their fine-tuned models, and evaluation scores on the test dataset.

\paragraph{Advanced Research:} Enhance the quality of your model through experimenting with advanced approaches, including creating synthetic data \citep{Lam2022-Speech-Synthetic,Moslem2024-IWSLT}, comparing end-to-end systems to cascaded systems \citep{Agarwal2023-IWSLT}, using language models (e.g. n-grams) \citep{Baevski2020-Wav2Vec}, domain adaptation \citep{Samarakoon2018-Speech-Domain-Adaptation}, or any other valid approach. By the end of the task, the participants should share their advanced models. They should also clarify how the advanced approach improved the speech translation quality compared to the original fine-tuned model.

\paragraph{Release \& Publication:} Write the project details to publish as a paper. Moreover, the outcomes of all the projects are publicly accessible.\footnote{\url{https://huggingface.co/SpeechT}}

\begin{figure}
    \centering
    \includegraphics[width=1.0\linewidth]{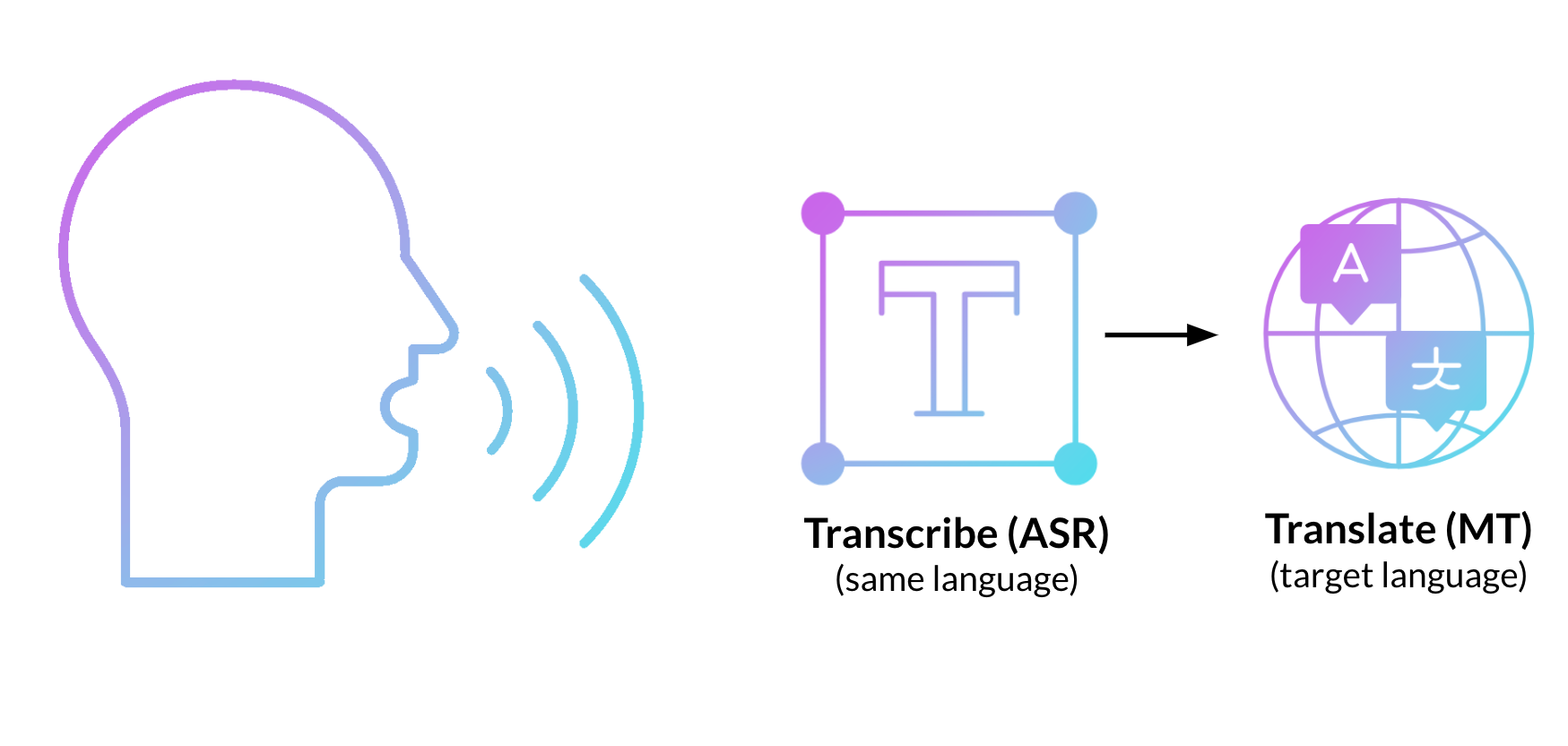}
    \caption{Cascaded Speech-to-Text System: Two models are trained, one for ASR, and one for MT of the transcriptions.}
    \label{fig:cascaded-speech}
\end{figure}

\begin{figure}
    \centering
    \includegraphics[width=0.95\linewidth]{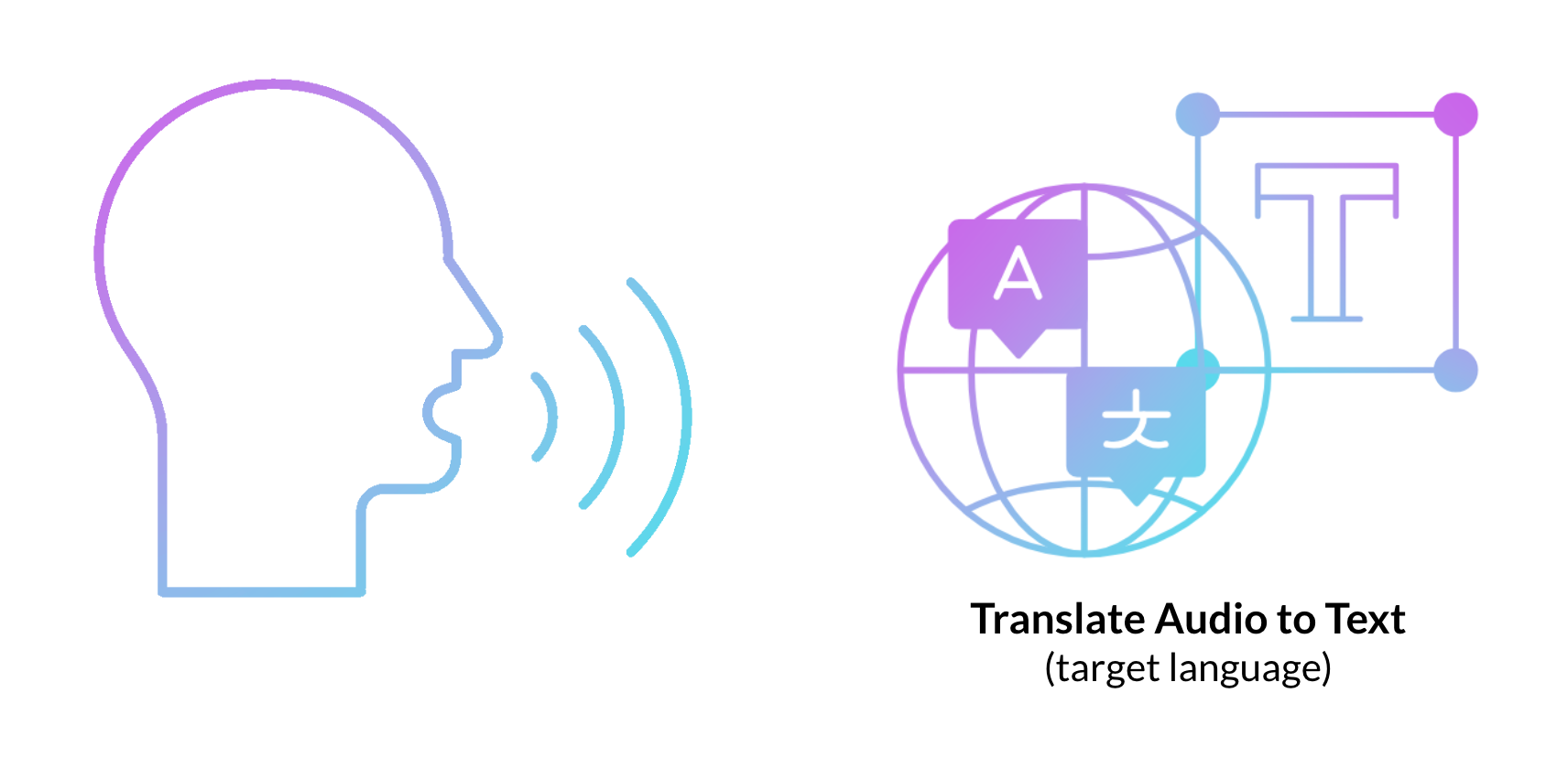}
    \caption{End-to-End Speech-to-Text System: One model is trained to generate the translation directly.}
    \label{fig:e2e-speech}
\end{figure}

\section{End-to-End vs. Cascaded systems}
\label{sec:e2e-vs-cascaded}

Speech translation systems can be (a) “cascaded” systems, or (b) “end-to-end” systems \citep{Agarwal2023-IWSLT,Ahmad2024-IWSLT}. Cascaded speech translation systems use two models, one for automatic speech recognition (ASR) and one for textual machine translation (MT) (cf. Figure~\ref{fig:cascaded-speech}). End-to-end speech translation systems use one model for the whole process (cf. Figure~\ref{fig:e2e-speech}).

\subsection{Cascaded Speech Translation}

Cascaded speech systems involve sequential modules for Automatic Speech Recognition (ASR), Machine Translation (MT), and optionally Text-to-Speech (TTS), simultaneously combined to deliver the output to the end user. The ASR system generates transcriptions from the input audio, and then the MT model translates the transcriptions into the target language. Among the advantages of building “cascaded” systems are:

\begin{itemize}
    \item Better quality in production.
    \item Each component (ASR, MT, TTS) can be individually optimized.
    \item Domain-specific (e.g. legal or medical) MT can be easily integrated.
\end{itemize}

\subsection{End-to-End (E2E) Speech Translation}

In end-to-end (E2E) speech systems, one model produces the whole process. E2E systems can also be extended with “cascaded” components. Among the advantages of building E2E systems are:

\begin{itemize}
    \item Simpler deployment
    \item Better performance (lower latency)
\end{itemize}

\section{Approaches to synthetic data}

When the data is limited for the language or domain, synthetic data can be used to augment the authentic data. Synthetic data for speech translation systems can be generated in diverse methods, including: 

\begin{itemize}
    \item Using TTS models to generate synthetic source audio for authentic translations \citep{Moslem2024-IWSLT}
    \item Using MT models to generate translations of audio transcriptions
    \item Sampling, translating, recombining: \citet{Lam2022-Speech-Synthetic} used an advanced approach to create synthetic data, by first chunking segments and transcriptions, creating a memory of prefix-suffix chunks based on part-of-speech tagging. Then they retrieve chunks from the memory to augment prefix chunks with similar suffix chunks. Finally, they translate the new transcription with MT. Tools such as WhisperX \citep{Bain2023-WhisperX} (based on Whisper) can be used for creating alignments based on word-level timestamps.
\end{itemize}

\begin{table*}[ht]
\centering
\begin{tabular}{@{}crrrl@{}}
\toprule
\textbf{Language Pair} & \textbf{Train} & \textbf{Dev} & \textbf{Test} & \textbf{Dataset}                                \\ \midrule
AR-EN                  & 2,228          & 278          & 279           & \textit{\small farahabdou/FLEURS-AR-EN-split}                   \\
BN-EN                  & 41,984          & 9,000          & 1,000           & \textit{\small satarupa22/indic-en-bn}                         \\
ES-JA                  & 9,972          & 1,440        & 1,345         & \textit{\small Marianoleiras/voxpopuli\_es-ja}                  \\
GL-EN                  & 4,798          & 507          & 282           & \textit{\small juanjucm/OpenHQ-SpeechT-GL-EN}                   \\
GL-EN                  & 2,742          & 496          & 212           & \textit{\small juanjucm/FLEURS-SpeechT-GL-EN}                   \\
ID-EN                  & 1,243          & 792          & 844           & \textit{\small cobrayyxx/COVOST2\_ID-EN} \\ \bottomrule
\end{tabular}
\caption{Data Statistics}
\label{tab:data}
\end{table*}

\section{Projects}

Most of the projects used a mix of data augmentation of authentic data with synthetic data, fine-tuning models, and comparing the performance of “end-to-end” speech systems to “cascaded” systems (cf. Section \ref{sec:e2e-vs-cascaded}). 

Participants used the Hugging Face Transformers library to fine-tune pretrained models. They fine-tuned Whisper \citep{Radford2022-Whisper} for “end-to-end” speech translation, and for ASR in the “cascaded” system. Moreover, they fine-tuned NLLB-200 \citep{NLLB2022} for text-to-text translation as part of “cascaded” speech translation systems. For evaluation, they used the sacreBLEU library \citep{Post2018-sacreBLEU} to obtain BLEU \citep{Papineni2002-BLEU} and ChrF++ \citep{Popovic2017-chrF++} scores. In addition, one of the participants calculated COMET scores \citep{Rei2020-COMET}. For inference, they either used the Transformers library or Faster-Whisper (based on CTranslate2 \citep{Klein2020-Efficient}) for audio translation and transcription with Whisper. For text-to-text translation with OPUS and NLLB-200 models, some of them used the Transformer library directly while others used CTranslate2 with \textit{float16} quantization, which is more efficient. For synthetic data generation, they used ChatGPT \citep{OpenAI2023-GPT-4} and OPUS \citep{Tiedemann2020-OPUS-MT} models.

Given that each participant chose a language pair, we dedicate a section for each project based on the language pair, including Galician-to-English, Indonesian-to-English, Spanish-to-Japanese, Arabic-to-English, and Bengali-to-English. Each language section describes data, modelling, and evaluation of each project.

\subsection{Galician-to-English}

\subsubsection{Data [GL-EN]}
\label{sec:data-galician}

In this project, two different Galician-to-English Speech Translation datasets have been curated. First, we compiled the dataset \textit{OpenHQ-SpeechT-GL-EN } from the \textit{crowdsourced high-quality Galician speech data set} by \citet{Kjartansson2020-Galician-Speech-Dataset}. After deduplicating the Galician audio-transcription pairs, we have applied a machine translation step to generate the corresponding English translations. More specifically, we have used GPT-4o \citep{Brown2020-GPT-3,OpenAI2023-GPT-4} with the following prompt:

\begin{itemize}
    \item [] [\{"role":"system", "content": "You are a helpful assistant that translates Galician (gl-ES) to English (en-XX).", \},
    \item [] \{"role": "user", "content": \{source\_text\}\}]
\end{itemize}

Given the absence of reference translation, we assessed the translation quality using CometKiwi (\textit{wmt23-cometkiwi-da-xl}) \citep{Rei2023-CometKiwi}, measuring an average score of 0.75. In total, this dataset contains approximately ten hours and twenty minutes of audio.

The second dataset is \textit{FLEURS-SpeechT-GL-EN}. This is a subset of the \textit{FLEURS} \citep{Conneau2023-FLEURS} dataset, which contains two thousand parallel audio-transcription pairs in a hundred and two languages. For assembling our dataset, each Galician audio-transcription pair has been aligned with the corresponding English text. For this dataset, we used the same method for measuring translation quality, achieving an average score of 0.76. After cleaning and deduplication, this dataset contains around ten hours of audio. Table \ref{tab:data} shows more details about the data.

\subsubsection{Modelling [GL-EN]}

We first employed Whisper to train an “end-to-end” speech translation system.
Whisper is a set of strong automatic speech recognition (ASR) architectures, trained on multilingual and multitask audio data. They can be further fine-tuned for speech translation. It supports Galician audio and text, making it a good choice for our data. Given our compute limitations, we experimented with two different backbones: \textit{whisper-small} and \textit{whisper-large-v3-turbo}, a simplified architecture of \textit{whisper-large} with fewer parameters in the decoder section. We fine-tuned both models over our two datasets (cf. Section \ref{sec:data-galician}).

To further improve our “end-to-end” results, we trained a “cascaded” system which splits the speech translation task into two consecutive steps (cf. Section~\ref{sec:e2e-vs-cascaded}). Intuitively, this separation allows each model to specialise in a specific step of the pipeline, while adding one extra level of explainability to the whole process. The first module consists of a \textit{whisper-large-v3-turbo}, this time in transcription mode, for generating Galician text given the input audio. Thereafter, on the same train split, we fine-tuned the MT model \textit{NLLB-200-distilled-600M} on Galician-to-English text translation.

Inference was performed using the Transformers library. More specifically, we used its pipeline functionality to encapsulate pre-processing and post-processing steps. Training and inference were run on one RTX 4090 GPU.

\subsubsection{Evaluation [GL-EN]}

For the \textit{FLEURS-SpeechT-GL-EN} dataset, the most performant “end-to-end” system was based on \textit{whisper-small}, achieving a BLEU score of 22.62 and a ChrF++ score of 46.11. For the \textit{OpenHQ-SpeechT-GL-EN} dataset, \textit{whisper-large-v3-turbo} was better, with a BLEU score of 55.65 and a ChrF++ score of 72.19. Regarding our cascaded system for \textit{FLEURS-SpeechT-GL-EN}, after using the MT model to translate the transcription generated by the ASR model, we obtained a BLEU score of 37.19 and a ChrF++ score of 61.33. For \textit{OpenHQ-SpeechT-GL-EN}, the cascaded approach resulted in a BLEU score of 66.05 and a ChrF++ score of 79.58. Hence, the cascaded approach, despite being more computationally demanding, allows for a better specialization for each part of the system, hence generating significantly better results (cf. Table \ref{tab:results}).

\subsection{Indonesian-to-English}

\subsubsection{Data [ID-EN]}

The dataset was compiled by extracting the English and Indonesian datasets from CoVoST2 \citep{Wang2021-CoVoST2}, a speech dataset in 21 languages, including Indonesian. Columns besides the index, Indonesian audio with its transcription, and English transcription were removed. The next preprocessing step was checking duplicate indices within each split and identifying overlapping indices across the splits. This dataset was first used to train an “end-to-end” speech-translation system. For speech translation using a “cascaded” system, two models were trained: an automatic speech recognition (ASR) model and a machine translation (MT) model. Hence, the audio and transcription columns were used to train the ASR model, while textual source and target columns were used to train the text-to-text MT model.

\subsubsection{Modelling [ID-EN]}

We employed different approaches for the speech-translation tasks, an “end-to-end” system and a “cascaded” system (cf. Section \ref{sec:e2e-vs-cascaded}). The pretrained model \textit{whisper-small} was used for training the “end-to-end” system. We fine-tuned the model with the Indonesian audio and English transcription directly. Meanwhile, in the “cascaded” system, the model was fine-tuned to predict the audio transcription in the same language, which is Indonesian. As a “cascaded” system requires an MT model for translating Indonesian transcription into English, we fine-tuned \textit{nllb-200-distilled-600M}, with batch size of 2 and gradient accumulation steps of 8 to simulate the effect of larger batch sizes. The model was trained for 10 epochs, saving the best epoch in the end. 

For inference, we used Faster-Whisper for both translation and transcription with Whisper after converting the model into the CTranslate2 formate with float16 quantization, with a batch size 5 and the VAD filter enabled.\footnote{Voice Audio Detection (VAD) removes low-amplitude samples
from an audio signal, which might represent silence or noise.} Similarly, for textual translation with NLLB-200, we used CTranslate2 with float16 quantization. Training was run on the T4 GPU from Google Colab, while inference used an RTX 2000 Ada GPU.

\subsubsection{Evaluation [ID-EN]}

The evaluation result of the “cascaded” system outperforms the “end-to-end” system on the \textit{CoVoST2} test set. The “end-to-end” system achieved a BLEU score of 37.02 and ChrF++ score of 56.04 after fine-tuning Whisper Small, considerably improving the baseline (whose scores were BLEU 25.87 and ChrF++ 43.79). The “cascaded” system which fine-tuned both Whisper for transcription and NLLB-200 for translation achieved 48.60 BLEU score and 65.10 ChrF++ score, which outperforms both the baseline (BLEU 38.24 and ChrF++	56.88) and the fine-tuned end-to-end model (cf. Table \ref{tab:results}).

\subsection{Spanish-to-Japanese}

\subsubsection{Data [ES-JA]}
\label{sec:data-es-ja}

The foundational dataset is \textit{VoxPopuli} \citep{Wang2021-VoxPopuli}, from which we extracted audio and Spanish transcriptions. We generated Japanese translations using OPUS models \citep{Tiedemann2020-OPUS-MT}, initially translating from Spanish to English and then from English to Japanese. While multilingual options existed, this two-step approach was chosen due to the strong performance of high-resource language pairs.
Post-processing was necessary to refine the dataset. First, we removed blank spaces, which are not typical in Japanese writing, ensuring proper formatting and consistency. Then, we eliminated empty texts and employed quality estimation with a threshold of 0.7 to filter out low-quality translations, using the CometKiwi (\textit{wmt23-cometkiwi-da-xl}) model. This process helped maintain alignment between the audio, transcriptions, and translations, resulting in a final dataset of approximately 12.7k rows.
Regarding content, the dataset consists of European Parliament event recordings featuring various Spanish accents. As a result, models trained on this data are likely to perform better in similar parliamentary or formal discourse scenarios (cf. Table \ref{tab:data}).

\subsubsection{Modelling [ES-JA]}

We built two systems for the Spanish-to-Japanese (ES-JA) speech translation task, an “end-to-end” system and a “cascaded” system (cf. Section \ref{sec:e2e-vs-cascaded}). The backbone of the “end-to-end” model is \textit{whisper-small}, which has been trained on the ES-JA VoxPopuli dataset \ref{sec:data-es-ja}. This \textit{whisper-small} model has been fine-tuned specifically for direct speech-to-text translation, meaning that the Spanish audio is encoded and directly decoded into Japanese, without requiring any intermediate transcription step. This approach offers a simpler architecture and a lower computational cost, since only one model is used, training and inference are more efficient.

On the contrary, the “cascaded” approach involves two separate models, (i) the \textit{whisper-small} for transcribing Spanish audio into text, and (ii) the \textit{nllb-200-distilled-600M} for translating the transcribed Spanish text into Japanese. While this method is more resource-intensive, it allows independent optimization of each component.

For inference, both approaches process Spanish audio inputs into Japanese text output. In the “end-to-end” approach, the model directly translates Spanish speech into Japanese in a single step (only one model is executed, taking less time and resources). However, in the “cascaded” approach there is a sequential process: The output of the model that transcribes Spanish into text is the input to the model that translates Spanish into Japanese (Two models are used, making it possible to optimize each of them but using more resources), providing a higher quality in terms of translation quality metrics. For this, we used the Hugging Face Transformers library pipelines: “automatic-speech-recognition” and “translation”. As for infrastructure, we conducted both training and inference of the models on one RTX~4090 GPU.

\subsubsection{Evaluation [ES-JA]}

The evaluation of the Spanish-to-Japanese translation models reveals a performance gap between the “end-to-end” and “cascaded” approaches. The “end-to-end” model scores on the test split indicate room for improvement, achieving a BLEU score of 20.86, a ChrF++ score of 23.36, and a COMET score of 77.7. This suggests that while the translations maintain some coherence, they lack the precision and fluency. In contrast, the “cascaded” approach outperforms the “end-to-end” model across all metrics. This system reaches a BLEU score of 35.32, a ChrF++ score of 32.82, and a COMET score of 89.86, demonstrating superior lexical and syntactic alignment with reference translations (cf. Table \ref{tab:results}).

\begin{table*}[ht]
\centering
\begin{tabular}{@{}ccccccc@{}}
\toprule
\textbf{Language Pair} & \textbf{System} & \textbf{Model} & \textbf{Type} & \textbf{Dataset} & \textbf{BLEU} & \textbf{ChrF++} \\ \midrule
                                
\multirow{12}{*}{\textbf{GL-EN}} & End-to-End & Whisper Small       & Baseline & Fleurs    & 16.01 & 44.99  \\
                                & End-to-End & Whisper Large Turbo & Baseline & Fleurs    & 5.09 & 26.59  \\
                                & Cascaded   & + NLLB-200 600M     & Baseline & Fleurs    & 34.47 & 59.29 \\
                                \cmidrule(l){2-7}
                                
                                & End-to-End & Whisper Small       & \cellcolor{cyan!8}Fine-tuned & Fleurs    & 22.62          & 46.11          \\
                                & End-to-End & Whisper Large Turbo & \cellcolor{cyan!8}Fine-tuned & Fleurs    & 18.96          & 46.00          \\
                                & Cascaded   & + NLLB-200 600M     & \cellcolor{cyan!8}Fine-tuned & Fleurs    & \textbf{37.19} & \textbf{61.33} \\ \cmidrule(l){2-7}

                                & End-to-End & Whisper Small       & Baseline & OpenHQ    & 21.46 & 41.12 \\
                                & End-to-End & Whisper Large Turbo & Baseline & OpenHQ    & 3.38 & 21.82 \\
                                & Cascaded   & + NLLB-200 600M     & Baseline & OpenHQ    & 43.01 & 64.52 \\
                                \cmidrule(l){2-7}
                                
                                & End-to-End & Whisper Small       & \cellcolor{cyan!8}Fine-tuned & OpenHQ    & 50.96          & 69.24          \\
                                & End-to-End & Whisper Large Turbo & \cellcolor{cyan!8}Fine-tuned & OpenHQ    & 55.64          & 72.19          \\
                                & Cascaded   & + NLLB-200 600M     & \cellcolor{cyan!8}Fine-tuned & OpenHQ    & \textbf{66.05} & \textbf{79.58} \\ \midrule

\multirow{4}{*}{\textbf{ID-EN}} & End-to-End & Whisper Small       & Baseline & CoVoST2  &  25.87 & 43.79  \\
                                & Cascaded   & + NLLB-200 600M     & Baseline  & CoVoST2 &  38.24 & 56.88 \\
                                \cmidrule(l){2-7}
                                
                                & End-to-End & Whisper Small       & \cellcolor{cyan!8}Fine-tuned          &                                  CoVoST2    & 37.02	      &         56.04          \\
                                & Cascaded   & + NLLB-200 600M     & \cellcolor{cyan!8}Fine-tuned          & CoVoST2    &  \textbf{48.60}   &	\textbf{65.10} \\  \midrule

\multirow{4}{*}{\textbf{ES-JA}} & End-to-End & Whisper Small       & Baseline  & VoxPopuli &  0.48 & 3.18  \\
                                & Cascaded   & + NLLB-200 600M     & Baseline  & VoxPopuli & 21.34 & 23.21  \\ \cmidrule(l){2-7}
                                
                                & End-to-End & Whisper Small       & \cellcolor{cyan!8}Fine-tuned    & VoxPopuli & 20.86          & 23.36          \\
                                & Cascaded   & + NLLB-200 600M     & \cellcolor{cyan!8}Fine-tuned    & VoxPopuli & \textbf{35.32} & \textbf{32.82} \\ \midrule

\multirow{3}{*}{\textbf{AR-EN}} & End-to-End & Whisper Small       & Baseline & Fleurs     & 5.65 & 31.75 \\
                                \cmidrule(l){2-7}
                                
                                & End-to-End & Whisper Small       & \cellcolor{cyan!8}Fine-tuned & Fleurs    & 15.06          & 39.03          \\
                                & Cascaded   & + NLLB-200 600M     & Baseline   & Fleurs    & \textbf{24.38} & \textbf{51.79} \\ \midrule
                                
\multirow{3}{*}{\textbf{BN-EN}} & End-to-End & Whisper Small       & Baseline & IndicVoices  & 6.33 & 24.60 \\
                                \cmidrule(l){2-7}
                                
                                & End-to-End & Whisper Small       & \cellcolor{cyan!8}Fine-tuned & IndicVoices    & 10.08 & 30.97          \\
                                & Cascaded   & + NLLB-200 600M           & Baseline   & IndicVoices    & \textbf{20.42} & \textbf{42.51} \\

                                \bottomrule
\end{tabular}
\caption{Results: Cascaded systems outperform end-to-end systems in speech translation across all language pairs.}
\label{tab:results}
\end{table*}

\subsection{Arabic-to-English \& Bengali-to-English}

Due to the similarity of the projects of the Arabic-to-English and Bengali-to-English language pairs, we combine them in one section. Unlike the aforementioned projects that fine-tuned models for all systems, these two projects fine-tuned models for the “end-to-end” system. In addition, the Bengali-to-English project fine-tuned Whisper for the “cascaded” system. However, both project used the baseline of NLLB-200 600M without fine-tuning.

\subsubsection{Data [AR-EN \& BN-EN]}

The dataset used in the Arabic-to-English project is a subset of the FLEURS dataset \citep{Conneau2023-FLEURS}, while the Bengali-to-English project used the IndicVoices dataset after filtering out segments whose mining scores are less than 0.7 \citep{Jain2024-BhasaAnuvaad,Javed2024-IndicVoices}.

The data is split into training and test sets to facilitate model training and evaluation. As the datasets include both the transcriptions and translations, it is useful for “end-to-end” speech translation tasks, as well as “cascaded” systems that involve separate speech recognition and machine translation models. Table \ref{tab:data} illustrates more details about the used data.

\subsubsection{Modelling [AR-EN \& BN-EN]}

Two approaches were employed for the Arabic-to-English and Bengali-to-English translation tasks:

End-to-End Model: The model utilizes whisper-small model, which is a pre-trained speech-to-text model capable of handling “end-to-end” speech translation. This model directly translates Arabic or Bengali speech into English text without intermediate steps. While the Arabic model was fine-tuned on the FLEURS dataset, the Bengali models were fine-tuned with the IndicVoices dataset.

Cascaded Model: This approach combines two models: (i) Automatic Speech Recognition (ASR) using the Whisper model to transcribe Arabic speech into Arabic text, and (ii) Machine Translation (MT) using NLLB-200 to translate the transcribed Arabic or Bengali text into English. 

For Arabic-to-English inference, the Hugging Face Transformers library was used for both speech-to-text transcription and text translation tasks, as well as “end-to-end” speech translation. For Bengali-to-English “end-to-end” translation, the Faster-Whisper library (based on CTranslate2) was used after converting the model with float16 quantization, while translation with NLLB-200 600M used CTranslate2. Training and inference utilized Google Colab, as well as GPU P100 on Kaggle and a multi-GPU setup comprising two NVIDIA T4 GPUs on Kaggle.

\subsubsection{Evaluation [AR-EN \& BN-EN]}

As in the case of other projects, the results of English-to-Arabic and Bengali-to-English speech translation indicate that the “cascaded” model outperforms the “end-to-end” model in terms of translation quality (cf. Table \ref{tab:results}).

\section{Conclusions}

The SpeechT mentorship brought together several practitioners and students from diverse companies and institutions across the world to explore speech translation. The participants have diverse backgrounds, ranging from generic software knowledge to text-to-text MT experience. Ultimately, five participants have made successful submissions and contributed to this work (cf. Section \ref{sec:contributions}).

Successful submissions incorporated a range of techniques. In particular, participants experimented with synthetic data generation with large language models (e.g. GPT4) and MT models (e.g. OPUS). The focus of most of the experiments was comparing the speech translation performance of “end-to-end” systems with “cascaded” systems (cf. Section \ref{sec:e2e-vs-cascaded}). For this purpose, the participants fine-tuned pretrained models, including Whisper and NLLB-200. While the “end-to-end” systems fine-tuned Whisper for direct speech translation, building the “cascaded” systems involved two steps, namely fine-tuning Whisper for ASR, and then employing an MT model (e.g. NLLB) for translation of the generated transcription. As Table~\ref{tab:results} illustrates, “cascaded” systems outperformed “end-to-end” across all language pairs. In conclusion, this mentorship has enabled the participants to experiment with various system designs and fine-tuning strategies, deepening their understanding of the speech translation area through hands-on practice.

\section{Contributions}
\label{sec:contributions}

\begin{small}
\begin{itemize}
    \item \textbf{Yasmin Moslem:} Organizer and mentor of \textit{SpeechT} mentorship in Speech Translation
\end{itemize}

\paragraph{Participants} \hspace{-1em} (alphabetically ordered)
\begin{itemize}
\setlength\itemsep{-0.05pt}
    \item \textbf{Farah Abdou:} Participant, Arabic-to-English Speech Translation
    \item \textbf{Juan Julián Cea Morán:} Participant, Galician-to-English Speech Translation
    \item \textbf{Mariano Gonzalez-Gomez:} Participant, Spanish-to-Japanese Speech Translation
    \item \textbf{Muhammad Hazim Al Farouq:} Participant, Indonesian-to-English Speech Translation
    \item \textbf{Satarupa Deb:} Participant, Bengali-to-English Speech Translation
\end{itemize}
\end{small}

\bibliography{paperpile}

\end{document}